\documentclass[wcp]{jmlr}

\usepackage{longtable}

\usepackage{booktabs}

\jmlrvolume{80}
\jmlryear{2020}
\jmlrworkshop{2020}

\title[Short Title]{IGLOO: Slicing the Feature Space to Represent Sequences}

  \author{\Name{Vsevolod Sourkov} \Email{vsourkov@gmail.com}\\
  \addr ReDNA Labs
  \AND
  \Name{TBA TBA} \Email{tba@test.com}\\
  \addr ReDNA Labs
 }

\editors{TBA}

\begin{document}

\maketitle

\begin{abstract}
Historically, Recurrent neural networks (RNNs) and its variants such as LSTM and GRU and more recently Transformers have been the standard go-to components when processing sequential data with neural networks. One notable issue is the relative difficulty to deal with long sequences (i.e. more than 20,000 steps). We introduce IGLOO, a new neural network architecture which aims at being efficient for short sequences but also at being able to deal with long sequences. IGLOO’s core idea is to use the relationships between non-local patches sliced out of the features maps of successively applied convolutions to build a representation for the sequence. We show that the model can deal with dependencies of more than 20,000 steps in a reasonable time frame. We stress test IGLOO on the copy-memory and addition tasks, as well as permuted MNIST (98.4\%). For a larger task we apply this new structure to the Wikitext-2 dataset ~\cite{merity2017pointer} and achieve a perplexity in line with baseline Transformers but lower than baseline AWD-LSTM. We also present how IGLOO is already used today in production for bioinformatics tasks.
\end{abstract}

\section{Introduction}
Until a few years ago RNNs were deemed to be the reference structure to use in a neural network as soon as there was a notion of sequence. Beyond the well known Long short-term memory (LSTM) ~\cite{hochreiter1997long} and the gated recurrent unit (GRU) ~\cite{cho2014learning} a plethora of other variations have been studied~\cite{jozefowicz2015empirical}, with no candidate showing a clear advantage according to those studies. Dealing with very long term dependencies is a current area of research and recent papers have introduced new variations which aim at fixing this issue and improve on the historical models: IndRNN ~\cite{li2018independently}, RNN with auxiliary losses:~\cite{trinh2018learning}. Earlier works also include the uRNN~\cite{arjovsky2016unitary}, Quasi-Recurrent Neural Networks (Q-RNN)~\cite{bradbury2016quasi}, Dilated RNN~\cite{chang2017dilated}, Recurrent additive networks~\cite{lee2017recurrent}, ChronoNet~\cite{roy2018chrononet}, EUNN ~\cite{jing2016tunable}, Kronecker Recurrent Units (KRU)~\cite{jose2017kronecker} and Recurrent Weight Average (RWA)~\cite{ostmeyer2017machine}. LWD-LSTM ~\cite{merity2018reg} is also a variation of the LSTM cell which has been extensively used and applied to the Wikitext-2 dataset.

Besides RNNs, convolutional networks have also been used to deal with sequences. Convincing results have been achieved in audio synthesis by~\cite{oord2016wavenet} using dilated causal convolutions. More generally the discussion in~\cite{bai2018empirical} presents a general structure based on convolutions, the Temporal Convolutional Network (TCN). It aims at learning sequences by using layers of convolutions stacked together and using different dilatation factors to  allow cells in the higher regions to have a large receptive field, hence capturing information from the whole input space. Typically the earliest layer (the one closest to the input) does not have any dilatation, while subsequent layers have exponentially increasing dilatations factors.

Another approach to deal with sequences is the Transformer model proposed in ~\cite{vaswani2017attention} where an attention mechanism is used to find representations. This has achieved state-of-the-art results on number of NLP tasks. One drawback of that approach is that when applied to sequences of length L, it is O($L^2$) in both computational and memory complexity, which makes it difficult to train long sequences on a single GPU. The Reformer ~\cite{kitaev2020reformer} fixes some of those issues by adding reversible layers and using locality-sensitive hashing for the attention computation. More recently, Synthesizer ~\cite{tay2020synthesizer} looks at generalizing the query, key, value paradigm of Transformers by replacing the dot product self-attention with alternative operations. 

Along with RNNs, Transformers and its variants and the TCN, this paper present a fourth class of neural nets which forms representations for sequences by using relationships between patches sliced out of the features map. Intuitively, it is using a form of correlation between different (non-local) parts of the sequence. Non-local neural networks focusing on 2D data have been explored in ~\cite{wang2017nonlocal}. This paper proposes IGLOO-base and IGLOO-seq, two new building blocks which yields competitive results on some common benchmarks as well as specific bioinformatics tasks.\newline

\textbf{Our Contributions} are as follows:
\begin{itemize}
\item We introduce IGLOO-base, a structure which finds a representation for a sequence by grouping together patches from different locations and processing them in a specific way. We show that this lightweight structure can be used to analyze sequences as long as 20,000 time steps. 
\end{itemize}
\begin{itemize}
\item We propose IGLOO-seq, a Transformer-like model but where the pairwise dot product attention mechanism is replaced with an IGLOO-base block. We show that this modification allows to reach competitive performance on selected benchmarks while having a computational requirements advantage in some circumstances.  
\end{itemize}

\section{IGLOO}
We examine two different cases: The first case produces a representation for the whole sequence which could then be sent to downstream layers. The second case produces a representation for each time step of the incoming sequence. This applies to NLP tasks such as Wikitext-2, where the next word at each time step should be predicted. The intuition behind the model is that the relationships between different time steps would allow to characterize the sequence as a whole. While the Transformer looks at all the possible pair-wise relationships in the input sequence using the dot product in the attention mechanism, our structure limits the number of relationships taken into consideration to a lower amount, given by the number of patches J.

\subsection{IGLOO-base}
This base model accepts an input X $\in R^{L\times M}$ and produces an output U $\in R^{J}$ where J is the number of patches processed in parallel and L is the length of the sequence. Using a causal 1D convolution (so that at time T, only data up to time T is available) with K filters, each time step can be represented as a vector $\in R^K$. Each element in this vector represents the activation for each filters for this time step. We apply such a convolution to X to obtain a full feature map F $\in R^{L\times K}$. 

\begin{figure}
  \centering
  \includegraphics[width=8cm]{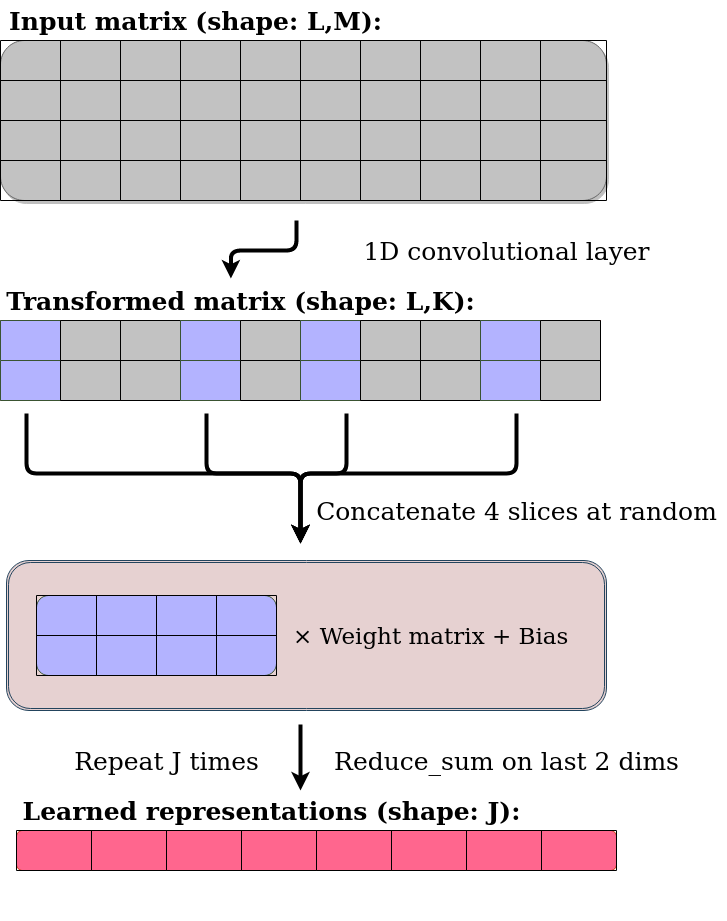}
  \caption{IGLOO-base architecture.}
\end{figure}

The IGLOO-base structure uses the gather\_nd operator on F to group p patches plucked from the time axis, out of the total L features available. A typical value of p is 4, and can be tuned given the problem to solve and the overall degree of over-fitting. Those patches are then stacked to obtain a matrix $\in R^{ p\times K}$. This operation is done in parallel J times to produce a matrix $\in R^{ p\times K\times J}$. This matrix is then multiplied point-wise by a trainable filter of the same size. One way to look at this operation is that the filter learns relationships between non-contiguous slices of F. We then add up together each elements resulting from the point-wise multiplication on the first and second axis to obtain a vector $\in R^{J}$. A trainable bias is also added and a non-linearity may be applied. As a result we obtain a vector U $\in R^{J}$ which represents the incoming sequence and can be fed to a downstream task. In total, we train O(J.K.p + J) parameters (not including the initial convolution C1). Note that this is independent of the length L of the sequence. 
Some remarks:
\begin{itemize}
\item The locations of the patches to be grouped can be decided randomly or structured in a specific deterministic way. In most cases either approach works. For language modelling tasks, it it beneficial to concentrate the patches around the current word of the input sequence since usually the last and recent words contribute most to the prediction of the following word. One possible issue is that the random patches do not cover a specific location absolutely necessary for the correct representation of that sequence. While this is possible, the fact that several consecutive layers of IGLOO are used mitigates the risk of this "catastrophic omitting". Using a larger size for the convolutional kernel may also reduce that risk. In practice this issue has not come up in our experiments.

\end{itemize}
\begin{itemize}
\item It can be seen that the number of parameters does not depend on L. From our experiments a value of J $<<$ L/3 gives good results. Besides, it is not required to calculate the full self-attention matrix as for Transformers in order to find a representation for the sequence, saving on the memory costly dot product attention operation.
\end{itemize}
\begin{itemize}
\item While traditional CNNs rely on network depth to bring together information from far away parts of the input, IGLOO-base directly samples patches from far away parts of the input so that it does not require depth per se. One layer can be enough in some cases. Nevertheless, since CNNs offer different levels of granularity at each consecutive layer, it can be interesting to use patches from consecutive layers too. We implement this by applying a different IGLOO-base to 1d convolutional layers applied consecutively to the input sequence. Assuming a depth of d, stacking the resulting IGLOO-base layers results in an output of size $\in R^{d\times J}$
\end{itemize}

\subsection{IGLOO-seq}

This model follows a Transformers-like architecture but without using self-attention. It accepts an input X $\in R^{L\times M}$ and produces an output U $\in R^{L\times K}$ where K is the last dimension of the "memory bank" B (as defined below) and L is the length of the sequence. This is typical of NLP tasks where the next word is to be predicted. Similarly to the Synthesizer ~\cite{tay2020synthesizer}, we also propose to replace the self-attention mechanism from Transformers (as illustrated in equation (1)) by a different block. The $Q.K^{T}$ matrix in equation (1) is of size $\in R^{L\times L}$. V is a linear projection of the input of size $\in R^{L\times M}$.

\begin{equation}
\ Attention(Q,K,V)= softmax(\frac{QK^{T}}{\sqrt{d}}).V
\end{equation}

\setlength{\parindent}{0ex}The Synthesizer proposes (among other variations) to replace the $Q.K^{T}$ operation with a two-layered feed forward layer, as in equation (2). 

\begin{equation} 
\ Synthesizer Dense(Q,V)= softmax(F(Q)).V
\end{equation}

where F(Q) is a feed forward layer of the form $F(Q)=W.(ReLU(W.Q+b))+b$. The F(Q) matrix is of size $\in R^{L\times L}$. V is a linear projection of the input of size $\in R^{L\times M}$. \newline

\begin{figure}
  \centering
  \includegraphics[width=16cm]{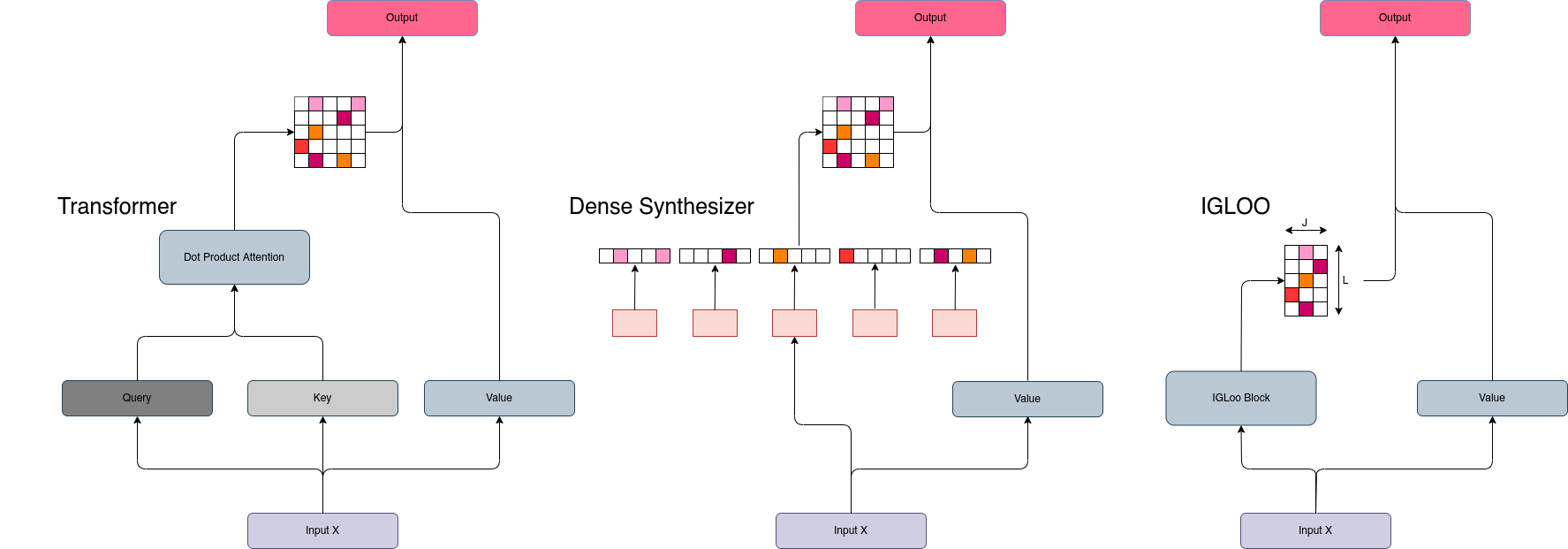}
  \caption{Comparison of different Transformer-like architectures.}
\end{figure}

In this paper, we propose to replace the $Q.K^{T}$ operation with an IGLOO-base block as shown in equation (3). Such a block generates in parallel a matrix U* $\in R^{L\times 1\times J}$ where J is the number of patches used for each time steps. Each L elements of U* is a representation of the sequence up to that point as obtained with IGLOO-base. We use a softmax operation over each representation to find the weights of the matrix which replaces the self-attention in the basic Transformer.

\begin{equation}
\ IGLOO seq(Q,V)= softmax(IGLOObase(Q)).V
\end{equation}\newline

One difference with Transformers and Synthesizers is that the "attention" matrix is not square. As such, the V part of the equation is expressed as:

\begin{equation}
\ V= H_J(F.W)\otimes B 
\end{equation}\newline
V represents the operations applied to the feature space F obtained from the initial convolution operation applied to the input X. Firstly, a projection into a space of size Z is applied to F by using a dot product with some weight W. At that point, $F.W$ $\in R^{L\times Z}$.
Then $H_J$, a tiling function which duplicates J times is applied, i.e. $ R^{L\times Z}$ $\rightarrow$ $ R^{L\times J\times Z}$. B is a trainable parameter of size $\in R^{L\times 1\times Z}$ and $\otimes$ is the point-wise multiplication operator. Taking the dot product between the softmax(U*) and V yields a representation for the sequence of shape $\in R^{L\times K}$. This amounts to taking a weighted average of the columns in V, with the weights being provided by IGLOO-base. It is possible to have a mixture of different IGLOO blocks each contributing to create its own output, then those outputs can be merged with an add operation. This allows to use different parameters for each block and therefore look at the sequence from different perspectives. For k such blocks, this would be expressed as:

\begin{equation}
\ IGLOO seq(Q,V) = \sum_{n=1}^{k} softmax(IGLOObase_n(Q)).V
\end{equation}\newline

\section{Discussion}

\subsection{Benefits compared to Transformers and Synthetizers}
The first benefit is the memory requirement in some conditions. Compared to $O(L\times L)$ for the Transformers-like structures, IGLOO-seq uses $O(L\times J\times K)$, which can be an advantage for large values of L and when J $<<$ L which is typically the case. For Tasks where only a representation for the sequence is required, the advantage of IGLOO-base is even more significant, since it requires only O(J.K.p + J)\newline

The second benefit is that compared to RNNs, IGLOO-base looks at the sequence as a whole and using the patching mechanism it does not suffer from the vanishing gradient issue which allows to deal with long sequences easily. One point to note is that the fixed sequence length must be decided before training and cannot be changed at inference time, just like with Transformers and contrary to RNNs. \newline

The third benefit of using a IGLOO-base block as proposed here to replace the Transformers and Synthetizers forms of attention is that doing so makes it unnecessary to use masking for language modelling tasks. Indeed, in order to avoid having information from subsequent words leaking to the current prediction via the softmax operation, it is usual to mask subsequent softmax values. As such, early tokens get contributions through the dot product of the attention and V only to the extent of their position in the sequence. For example, the prediction for the subsequent word at the N-th position will get contribution from the softmax only for the first N elements. When N is small, only a few elements are eventually used to form the prediction. In the case of IGLOO-seq, a constant J elements are contributing in the softmax dot product, so that early elements are not under-fitted by having fewer parameters to work with compared to later elements.

\subsection{Adding Depth}
Previously we have only considered gathering p patches over the initial convolutional layer, but we find that it is useful to repeat the operation for successive convolutional layers. Intuitively it allows to look at the sequence at different levels of granularity and also increases the recipient field of IGLOO-base. In practice using stacks of 2 or 3 layers works well.

\subsection{Residual Connections}
Just as for Transformers, in the case of IGLOO-seq we use a residual connection between the input X, and the output from the IGLOO-seq structure. Also like for Transformers, a double layered feed forward layer is added at each level of depth to increase the expressive power of the structure. We note that contrary to Transformers, positional encodings are not required, since the locations of the patches are explicitly generated while initializing the network.

\section{Applications}
\subsection{Copy Memory task}
\textbf{Dataset} This task was initially introduced in~\cite{hochreiter1997long}. We are given a vector of size T+20, where the first 10 elements G=[G0, G1, .., G9] are random integers between 1 and 8, the next T-1 elements are 0’s, the following element is a 9 (which serves as a marker), and the following 10 elements are 0’s again. The task is to generate a sequence which is zero everywhere except for the last 10 items which should be able to reproduce G. The task A can be considered a classification problem and the loss used will be categorical cross entropy (across 8 categories). The model will output 8 classes which will be compared to the actual classes. 

\textbf{Setup} The metric we keep track of is the wall time needed to reach a certain accuracy on the test set. This makes more sense than comparing number of epochs required to reach those targets since epochs have a different durations for those various structures. We use a Batch size of 128 for all models. We use 128 hidden states for Vanilla GRU, Vanilla LSTM, CuDNNGRU and CuDNNLSTM. Models may differ in terms of parameters size but we are mostly interested in convergence speed in those experiments. Experiments are run in Keras with Tensorflow backend on GTX1060 GPU. For IGLOO, we use K=5 and causal padding along with varying number of patches L. All values are averaged over 10 runs.

\begin{table}
  \caption{\textbf{Copy Memory T=30}. Time to reach accuracy $>$ 0.99}
  \label{sample-table}
  \centering
  \begin{tabular}{lll}
    \toprule
    Model (Hidden)     & Time (s)     & Params \\
    \midrule
    CuDNNGRU (128) &  1000+  & 51K     \\
    CuDNNLSTM (128) & 1000+  & 68K     \\
    indRNN (2X128) & 1000+  & 51K     \\
    RWA (168) & 1000+  & 58K     \\
    Transformer  & 498  & 29K     \\
    TCN (6X16) & 21  & 70K     \\
    IGLOO (100) & \textbf{62}  & 59K     \\
    \bottomrule
  \end{tabular}
\end{table}

\begin{table}
  \caption{\textbf{Copy Memory - IGLOO model}. Time to reach accuracy $>$ 0.99}
  \label{sample-table}
  \centering
  \begin{tabular}{llll}
    \toprule
    Model(size)  & Time steps     & Time (s)     & Params \\
    \midrule
    IGLOO-seq(200 patches) & T=100  & 57  & 227K     \\
    Transformer(depth 3) & T=100  & $>$ 1000  & 210K     \\
    IGLOO-seq(500 patches/2 stacks) &  T=1,000  & 118  & 258K     \\
    Transformer(depth 3) & T=1000  & $>$ 1000  & 368K     \\
    IGLOO-seq(4000 patches/2 stacks) & T=10,000 & 2224  & 21M     \\

    \bottomrule
  \end{tabular}
\end{table}

\textbf{Results} As can be seen in Table 1, for a small value of T, we find that the optimized version of the GRU cell converges but very slowly. The vanilla version is even slower. The Transformer takes close to 500s to converge and needs an initial Conv1D layer in order to increase the dimensionality of the input sequence. Other cells except the TCN also take a long time to converge. Above 1000 steps the TCN sometimes gets stuck in a local minimum. From Table 2 it appears that for more than 1,000 steps, IGLOO-seq is an order of magnitude faster for this task. There is still convergence for up to \textbf{25,000 steps} which to our knowledge is longer than for any other structure in the literature. After 20,000 steps convergence can still happen in some cases. To the best of our knowledge no other method in the literature achieves convergence for that many steps on this benchmark.

\begin{table}
  \caption{\textbf{Addition task T=200}. Time to reach loss $<$ 0.01}
  \label{sample-table}
  \centering
  \begin{tabular}{lll}
    \toprule
    Model (Hidden)     & Time (s)     & Params \\
    \midrule
    Vanilla GRU (128) & 153  & 50K     \\
    Vanilla LSTM (128) & $>$ 1500  & 67K     \\
    CuDNNGRU (128) & 24  & 50K     \\
     CuDNNLSTM (128) & 99  & 67K     \\  
      IndRNN (2*128) & 401  & 50K     \\
      RWA (164) & 235  & 55K     \\ 
      TCN (18) & 82  & 46K     \\ 
  IGLOO (L=500) & \textbf{8}  & 11K     \\    
    \bottomrule
  \end{tabular}
\end{table}

\begin{table}
  \caption{\textbf{Addition task T=1000}. Time to reach loss $<$ 0.01}
  \label{sample-table}
  \centering
  \begin{tabular}{lll}
    \toprule
    Model (Hidden)     & Time (s)     & Params \\
    \midrule
    Vanilla GRU (226) & 1298  & 155K     \\
    Vanilla LSTM (196) &  $>$ 2000  & 155K     \\
    CuDNNGRU (226) & 903  & 156K     \\
     CuDNNLSTM (196) &  $>$ 2000  & 156K     \\  
      IndRNN (2*226) &  $>$ 2000  & 160K     \\
      RWA (276) & 442  & 155K     \\ 
      TCN 24) & 171  & 142K     \\ 
  IGLOO (L=2000) & \textbf{17}  & 133K     \\    
    \bottomrule
  \end{tabular}
\end{table}

\subsection{Addition Task}
\textbf{Dataset} This task (Task B) also tests the structure capacity to deal with long term memory. Given an input vector of shape (2,T), with the first row being random numbers and the second row being 0 everywhere except at 2 random locations where it is 1, the purpose is to generate the result of adding the two numbers marked by the locations where there is a 1. This is a regression task and we shall investigate results for different size of T and different models. MSE is used as the loss function.

\textbf{Setup} We use 22,500 samples for the training, and 2,500 for the test set. For T=200, the IGLOO-base structure used has 500 patches. Since we only output the last time step rather than the whole sequence, the number of parameters is small compared to other structures. A learning rate of 0.005 is used throughout. Batch size is 100.  We also use L=5 and a patch size of 4. We use a causal padding style and a learning rate of 0.005. We use the Adam optimizer with clip norm =1. The hidden dimensions for other cells are shown in the results table. They are used so that the number of parameters is similar for all cells in the experiment. For T=1,000, We use an IGLOO cell with 2,000 patches, a patch size of 4, and 3 stacks. We use a causal padding style and a learning rate of 0.005. We use the Adam optimizer with clip norm =1. For IGLOO with 5,000 steps, we use J=5,000 patches, 3 stacks.

\textbf{Results} For T=200 (Table 3), the vanilla LSTM takes much longer than other structures to converge, whereas the CuDNNLSTM and CuDNNGRU cells show competitive convergence. The Igloo cell is the fastest in this experiment.
For 1000 time steps, some of the cells take more than 2000 seconds to converge and  the igloo cell is still the fastest. For 20,000 steps there is still convergence. To the best of our knowledge no other method in the literature achieves convergence for that many steps on this benchmark.

\begin{table}
  \caption{\textbf{MNIST - pMNIST}. Accuracy in \%}
  \label{sample-table}
  \centering
  \begin{tabular}{lll}
    \toprule
    Model (reference)     & MNIST     & pMNIST \\
    \midrule
    iRNN (Le et al., 2015) & 97.0  & 82.0    \\
   uRNN (Arjovsky et al., 2016) & 95.1  & 91.4     \\ 
      LSTM & 98.3  & 89.4     \\
      EURNN(Jing et al.,2016) & -  & 93.7     \\
   TCN (Bai et al., 2018) & 99.0  & 97.2     \\
r-LSTM (Trinh et al., 2018)        & 98.4  & 95.2     \\       
IndRNN (Li et al., 2018)        & 99.0  & 96.0     \\
KRU(Jose et al.,2017)        & 96.4  & 94.5     \\ 
Dilated GRU (Chang et al., 2018)       & \textbf{99.2}  & 94.6     \\ 
IGLOO - ours       &  98.6  & \textbf{98.4}     \\  
    \bottomrule
  \end{tabular}
\end{table}

\subsection{Sequential MNIST and permuted MNIST}
\textbf{Dataset} Another dataset to benchmark the ability of a network to learn long term dependencies is the sequential MNIST task. In this task pixels are fed into a recurrent model before making a prediction. Since the MNIST dataset consists of 28*28 pixels sized images, once flattened they essentially amount to 784-long sequences. A variation is to randomly permute the pixels (the same way across all samples). This breaks the local structure of the image and should result in more complexity for the task. We report the results from the literature and compare them to applying IGLOO to this task.

\textbf{Setup} We use results from previous works as reported in the notes and compare the results to the IGLOO-base structure. For all results except IGLOO-base we just report what has been claimed in terms of performance in those various papers. For the pMNIST we use J=2500 patches, a patch size of 4, a spatial dropout of 0.15, 4 stacks and an initial Conv1D convolution of size K=8. For MNIST we use J=2500 patches, a patch size of 4, a spatial dropout of 0.2, 6 stacks and an initial Conv1D convolution of size K=8. We train over 200 epochs with a batch size of 128.

\textbf{Results} It appears in Table 5 that IGLOO-base is competitive on the sequential MNIST dataset with a performance above the LSTM but slightly below the TCN and IndRNN (those performances are reported in the original papers and have not been reproduced here). For the pMNIST task, \textbf{IGLOO-base is performing the best across all reported results in the literature}. We see that the pMNIST and MNIST performances are very similar. IGLOO may be less impacted by permutation than RNN style structures because it is finding a representation for a sequence not by looking at each element sequentially but as a whole, taking patches from the whole input space.

We note that while IGLOO-base and CuDNN LSTM run at similar speed of 30 seconds per epoch, the LSTM is much slower and takes about 540 seconds per epoch for a 128 hidden layers cell. Therefore we achieve superior accuracy for the pMNIST benchmark with speed levels (per epoch) similar to the fast NVIDIA optimized CuDNN LSTM cell. IGLOO is also competitive in terms of pure wall time just as for previous experiments.

\begin{table}
  \caption{\textbf{Perplexities on the Wikitext-2 dataset.}}
  \label{sample-table}
  \centering
  \begin{tabular}{llll}
    \toprule
    Model (reference)  & Nb Params   & Validation     & Test \\
    \midrule
    Variational LSTM (Inan et al. 2017)  & 28M & 92.3  & 87.7    \\
    AWD-LSTM (Merity et al. 2017)  & 33M & 68.6  & 65.8   \\
    Transformer (Vaswani et al. 2017)  & 10M & 58.1  & 56.1    \\
    IGLOO-seq-100 (ours)  & 18M & 58.3  & 56.9    \\     
    IGLOO-seq-50 (ours)  & 17.2M & 58.6  & 57.1    \\ 
    \bottomrule
  \end{tabular}
\end{table}

\subsection{Wikitext-2 Dataset}

\setlength{\parindent}{0ex}\textbf{Dataset} Following ~\cite{wang2019improving} we use the Wikitext-2 dataset ~\cite{merity2017pointer} to compare IGLOO-seq to different models. This dataset has a total of 2M tokens in the training set and has been extensively benchmarked with many NLP models. As it is composed of full articles, the dataset is well suited for models that can take advantage of long term dependencies which is one important aspect we want to test in this section. Namely, we want to experiment test whether IGLOO-seq is able to capture long term dependencies by using the IGLOO-base block.

\textbf{Setup} We implement a baseline IGLOO-seq model without using any specific optimization relative to this task such as dynamic evaluation of partial shuffling. We use an IGLOO-base block made of 2 stack and a depth of 2. We use 100 and 50 as the number of patches J, noting that the dataset has 256 time steps. For the initial convolutional we use 256 filters, a kernel size of 3 and a causal padding. For this experiment we use a random procedure to generate the J patches locations for each step by sampling from a gaussian centered on that step index. We choose the variance of that gaussian distribution to allocate a larger number of patches in the vicinity of the current time step. Other options are possible and we leave it for further study. If no seed is used, then experiments should differ because of the difference in patch locations. However we notice a very small variance in the final predictions, somewhere around 0.5 perplexity points.

\textbf{Results}  Table 6 shows the performance for this task. The performance is in line with baseline Transformers. We note that different numbers of patches J (100 and 50) produce similar results, but have a different memory usage. The model using a smaller number of patches J can therefore use a larger batch size when trained on a single GPU.

\section{Discussion}

The first point we would like to address is whether it is possible to approximate Transformers with IGLOO-seq. It appears that if patches were selected in a deterministic way using a patch size of 2 and a number of patches J larger than or equal to the sequence length L, one would be able to replicate the pairwise dot-product found in the base Transformer. It should be noted that the representation for the Nth element, rather than having N elements contributing to it using the softmax dot-product in Transformers, would now have J elements, with J $>$ N. Therefore with this set-up IGLOO-seq gives a higher bound approximation of Transformers (from a perspective of the number of parameters).\newline 

The second point we would like to discuss is the usage of IGLOO-base in production. Since the performance of IGLOO-base is particularly suited to finding representations for long sequences, it seems particularly well suited for bioinformatics tasks. Having adopted this structure in their work: RNAsamba ~\cite{camargo2020rnasamba}, the authors use an IGLOO-base block to predict the coding potential of RNA modecules using their nucleotide sequence. From the sequence of nucleotides (A,C,G,T) the algorithm predicts whether it is protein coding or not. In their experiments, they analyze sequences of 3,000 elements (or time steps). They also use a number of patches J=600 for their expriment. This choice illustrates that J $<<$ L for a real world task and that the benefits enumerated above in this paper become apparent in terms of ressources usage.
Using The FEELnc dataset: ~\cite{wucher2017feelnc}, the authors of that paper obtain state of the art results with an accuracy of 98.46\% on a balanced classes test set, improving on the the previous methods using RNNs and Transformers. Other bionformatics tasks such as predicting circular RNA versus other long non coding RNA can also be processed with IGLOO-seq since again it involves looking at a large number of time-steps (i.e. nucleotides). 
It is promising to see that IGLOO has already been adopted in certain industries to respond to real world challenges and tasks. We are looking forward to getting it more widely used in the NLP area.

\section{Conclusion}

In this work we have presented two new neural network structures which can be used to find representations for sequences. We show that one of them is closely related to Transformers. Performance wise, they appear to be on par with Transformers and the memory usage makes them favourable for long sequences. Recent papers have been proposing applying Transformer-like structures to image data rather than just sequence data and this is also an area which we would like to investigate in forthcoming papers. A IGLOO-2D mechanism can be used which would gather patches from the 2D feature space extracted from images by a 2D convolutional layer. Given the possible memory advantage of IGLOO versus Transformers it could be of benefit when working with 2D data where memory usage is a current bottleneck for Transformers.
More generally, we find that using a form of correlation between portions of sequences or images is a promising avenue for future research. We also plan to explore more applications of IGLOO-base and IGLOO-seq in bioinformatics where it is already used to establish state-of-the-art results.

\nocite{*}

\acks{TBA}

\bibliography{acml18}

\appendix

\section{First Appendix}\label{apd:first}

\end{document}